\documentclass{article}
\usepackage{spconf,amsmath,epsfig}

\usepackage{hyperref}
\usepackage{amsmath,amssymb}
\usepackage{tikz}
\usepackage{pgfplots}
\usepackage{multirow}
\usepackage{makecell}
\usepackage{xcolor}

\let\OLDthebibliography\thebibliography
\renewcommand\thebibliography[1]{
  \OLDthebibliography{#1}
  \setlength{\parskip}{0pt}
  \setlength{\itemsep}{0pt plus 0.3ex}
}

\begin{document}\sloppy

\def\x{{\mathbf x}}
\def\L{{\cal L}}

\title{Body-Part Joint Detection and Association via Extended Object Representation}
%
\name{Anonymous ICME submission}
\address{}


\name{Huayi Zhou$^{\dagger}$ \qquad Fei Jiang$^{\ddag}$ \sthanks{* Corresponding author: fjiang@mail.ecnu.edu.cn.} \qquad Hongtao Lu$^{\dagger}$ \thanks{This paper is supported by NSFC (No. 62176155, 62207014), Shanghai Municipal Science and Technology Major Project (2021SHZDZX0102). Hongtao Lu is also with MOE Key Lab of Artificial Intelligence, AI Institute, Shanghai Jiao Tong University.}}
\address{$^{\dagger}$ Department of Computer Science and Engineering, Shanghai Jiao Tong University; \\
    $^{\ddag}$ Shanghai Institute of AI Education, East China Normal University}


\maketitle

\begin{abstract}
The detection of human body and its related parts (e.g., face, head or hands) have been intensively studied and greatly improved since the breakthrough of deep CNNs. However, most of these detectors are trained independently, making it a challenging task to associate detected body parts with people. This paper focuses on the problem of joint detection of human body and its corresponding parts. Specifically, we propose a novel extended object representation that integrates the center location offsets of body or its parts, and construct a dense single-stage anchor-based Body-Part Joint Detector (BPJDet). Body-part associations in BPJDet are embedded into the unified representation which contains both the semantic and geometric information. Therefore, BPJDet does not suffer from error-prone association post-matching, and has a better accuracy-speed trade-off. Furthermore, BPJDet can be seamlessly generalized to jointly detect any body part. To verify the effectiveness and superiority of our method, we conduct extensive experiments on the CityPersons, CrowdHuman and BodyHands datasets. The proposed BPJDet detector achieves state-of-the-art association performance on these three benchmarks while maintains high accuracy of detection. Code is in \url{https://github.com/hnuzhy/BPJDet}. 
\end{abstract}
\begin{keywords}
Association, body-part joint detection, object detection, object representation
\end{keywords}
\section{Introduction}

Human body detection \cite{zhang2018occlusion, wang2018repulsion, chu2020detection} is a research hotspot in computer vision. Accurate and fast human detection in an arbitrary scene can support many down-stream vision tasks such as pedestrian re-id, person tracking and human pose estimation. Also, the detection of body parts like face \cite{hu2017finding, deng2020retinaface}, head \cite{vu2015context, le2018key} and hands \cite{zhou2018raising, narasimhaswamy2022whose} is equally important. They may serve as precursors to specific missions like face recognition, crowd counting and hand pose estimation. Although the detection performance of human bodies and related parts has been significantly improved due to breakthroughs of deep CNN-based general object detection (e.g., Faster R-CNN \cite{ren2015faster}, FPN \cite{lin2017feature}, RetinaNet \cite{lin2017focal} and YOLO \cite{redmon2016you}) and construction of large-scale high-quality datasets (e.g., COCO \cite{lin2014microsoft}, CityPersons \cite{zhang2017citypersons} and CrowdHuman \cite{shao2018crowdhuman}), the joint discovery of human body and its parts is still challenging but meaningful.

\begin{figure}[!t]
	\centering
	\includegraphics[width=\columnwidth]{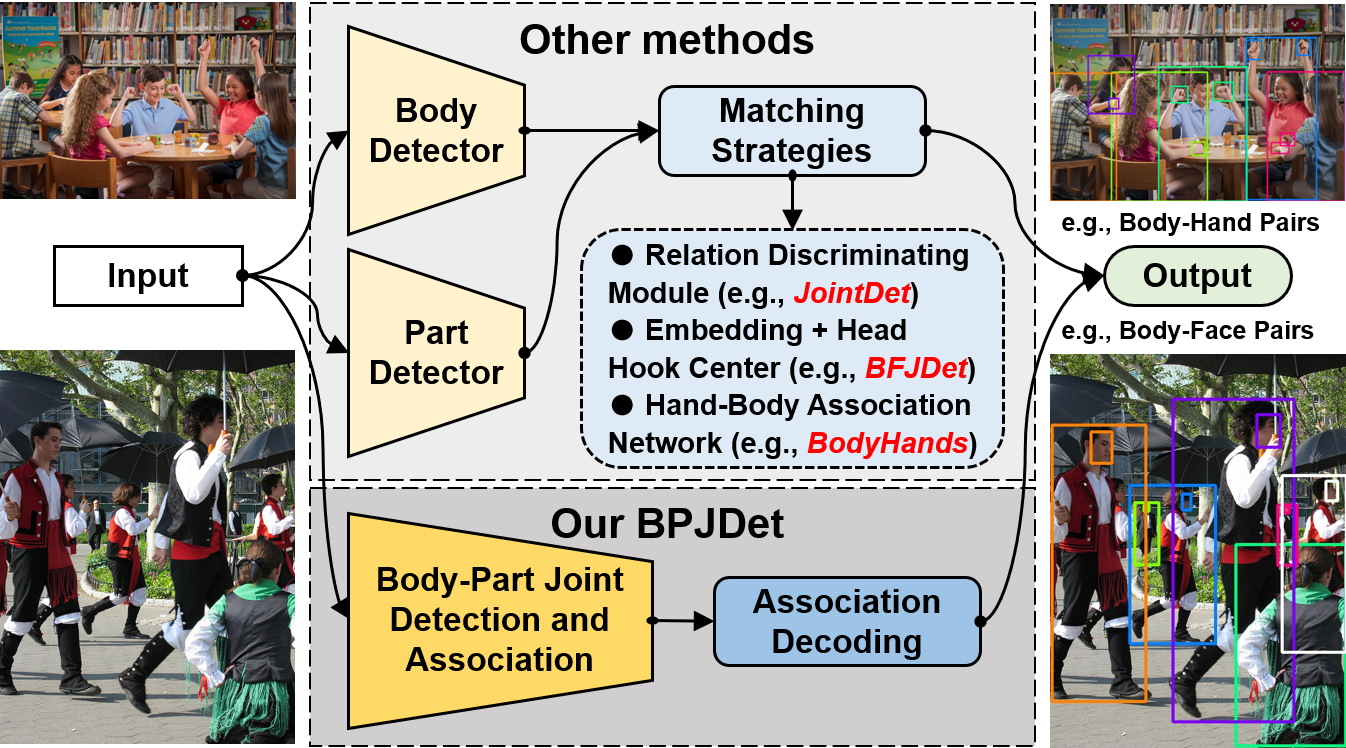}
	\caption{The illustration of the difference between our proposed single-stage BPJDet and other two-stage body-part joint detection methods (e.g., JointDet \cite{chi2020relational}, BFJDet \cite{wan2021body} and BodyHands \cite{narasimhaswamy2022whose}). Their two-stage refers to training the detection and association modules separately, rather than our joint detection and association framework.}
	\label{illustrations}
	\vspace{-10pt}
\end{figure}


In this paper, we study the problem of body-part joint detection and propose a Body-Part Joint Detector (BPJDet). As shown in Fig. \ref{illustrations}, the body part can be a face or hand. Our goal is to improve the body-part association accuracy on the premise of ensuring the detection precision. Prior to ours, several researches have attempted to address the joint body and part detection problem. DA-RCNN \cite{zhang2019double} and JointDet \cite{chi2020relational} focus on joint body-head detection. BFJDet \cite{wan2021body} concentrates on the similarity body-face joint detection task. BodyHands \cite{narasimhaswamy2022whose} and \cite{zhou2018raising} tackle the problem of detecting hands and finding the location of the corresponding person.

Unlike above methods using explicit strategies to model body-part relationships by exploiting heuristic post-matching or learning branched association networks, as illustrated in Fig. \ref{illustrations}, we propose a novel extended object representation that can be harmlessly applied to single-stage anchor-based detectors like YOLO series \cite{redmon2016you, redmon2018yolov3, jocher2020ultralytics}. Besides bounding boxes, confidence and objectness contained in the classic object representation, our extension adds location offsets of body parts. Regressing offsets is popular in object detection \cite{zhou2019objects, tian2019fcos} and human pose estimation \cite{cao2017realtime, nie2019single, mao2021fcpose, xiao2022adaptivepose}. Our simple yet efficient innovation has at least three advantages. 1) The relational learning of body-part can benefit from the training of general detectors without painstakingly designing association subnets. 2) The final prediction naturally implies the detected body bounding boxes and their associated parts, which avoids error-prone and tedious association post-processing. 3) This general representation enables us to jointly detect arbitrary body part such as face and hand without major modifications. In experiments, we performed extensive tests on three public datasets to verify the superiority of our method.

To sum up, we mainly have following three contributions: 1) We propose a novel end-to-end trainable Body-Part Joint Detector (BPJDet) by extending the classic object representation with regressing offsets of body parts. 2) We reveal the feasibility and expedience of joint training of bounding boxes and offsets with designing suitable multi-task loss functions. 3) We achieve state-of-the-art body-part association performance on three public benchmarks while maintaining high detection accuracy of body and parts.


\section{Related Work}

{\bf Human Body and Part Detection:} We here discuss emerging powerful CNN-based detectors that achieve promising results over traditional approaches using hand-crafted features. For human body detection, it belongs to either general object detection containing the person category \cite{ren2015faster, lin2017feature, lin2017focal, redmon2016you, redmon2018yolov3, zhou2019objects, tian2019fcos, jocher2020ultralytics} trained on common datasets like COCO \cite{lin2014microsoft}, or pedestrian detection \cite{zhang2018occlusion, wang2018repulsion, chu2020detection, liu2019adaptive, xu2020beta} trained on specific benchmarks CityPersons \cite{zhang2017citypersons} and CrowdHuman \cite{shao2018crowdhuman}. A key challenge in human body detection is occlusion. Therefore, many researches propose customized loss functions \cite{zhang2018occlusion, wang2018repulsion}, improved NMS \cite{liu2019adaptive, chu2020detection} or tailored distribution model \cite{xu2020beta} for alleviating problems of crowded people detection. On the other hand, detection of body part has also been intensively and extensively studied. Face detection \cite{hu2017finding, deng2020retinaface} and head detection \cite{vu2015context, le2018key} are two mostly vigorous fields. Face detectors are usually based on well-designed networks and trained on dedicated datasets. Head detectors are rarely studied alone, and often used for facilitating crowd counting, enhancing crowd human detection or preprocessing of head pose estimation. Besides, some literatures present detection of hands such as hand-raisers \cite{zhou2018raising} or body-hand pairs \cite{narasimhaswamy2022whose}.

{\bf Body-Part Joint Detection:} We mainly pay attention to the joint detection of face, head and hands body part. First of all, body-head joint detection is the most popular couple, because the human head is a salient structural body part and plays a vital role in recognizing people, especially when it is occluded. For example, DA-RCNN \cite{zhang2019double} proposes to handle the crowd occlusion problem in human detection by capturing and cross-optimizing body and head parts in pairs with double anchor RPN. JointDet \cite{chi2020relational} presents a head-body relationship discriminating module to perform relational learning between heads and human bodies. Recently, BFJDet \cite{wan2021body} investigates the performance of body-face joint detection for the first time, and proposes a bottom-up scheme that outputs body-face pairs for each pedestrian. It also adopts independent detection followed by pairwise association. For body-hand joint detection, \cite{zhou2018raising} devises heuristic  strategies to match hand-raising gestures with body skeletons \cite{cao2017realtime} in classroom scenes. BodyHands \cite{narasimhaswamy2022whose}, in unconstrained conditions, proposes a novel association network to jointly detect hands and the body location, and introduces a new corresponding hand-body association dataset. Unlike all of them, our approach BPJDet is not restricted to a certain body part, and tackles detection and association in an end-to-end way.

{\bf Representation of Object Detection:} Most modern object detection systems including anchor-based \cite{ren2015faster, lin2017feature, lin2017focal} and anchor-free \cite{redmon2016you, redmon2018yolov3, tian2019fcos, jocher2020ultralytics} detectors represent objects with bounding boxes. CenterNet \cite{zhou2019objects} models objects using heatmap-based center points and represents human poses as a 2K-dimensional property of the center point. It firstly extends the traditional object representation, and reveals essential overlap between the tasks of object detection and human pose estimation. Similarly, for instance, FCPose \cite{mao2021fcpose} adapts single-stage object detector FCOS \cite{tian2019fcos} with proposed dynamic filters to process person detections by predicting keypoint heatmaps and regressing offsets. Other single-stage pose estimators SPM \cite{nie2019single} and AdaptivePose \cite{xiao2022adaptivepose} also learn to regress keypoints as series of joint offsets. These works motivate us to detect human body with regressing its parts jointly by an extended object representation.

\section{Our Method}

\begin{figure*}[!t]
	\centering
	\includegraphics[width=0.98\textwidth]{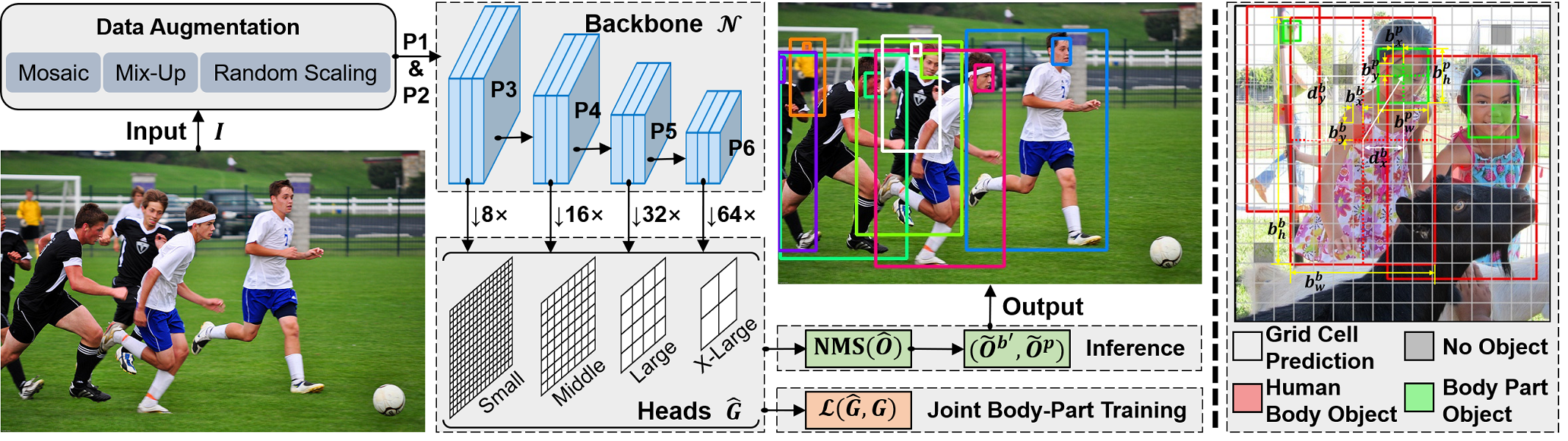}
	\vspace{-10pt}
	\caption{{\bf Left:} Our BPJDet adopts YOLOv5 as the backbone $\mathcal{N}$ to extract features and predict grids $\mathit{\widehat{G}}$ from one augmented input image $\mathbf{I}$. During training, target grids $\mathit{G}$ are used to supervise the loss function $\mathcal{L}$. In inference, NMS and association decoding are applied on predicted objects $\widehat{\mathbf{O}}$ to obtain final boxes set $\widetilde{\mathcal{O}}^{box}$ and related offsets set $\widetilde{\mathcal{O}}^{dis}$. {\bf Right:} Examples for grid cell predictions with human body objects in red color and body part objects (e.g., face) in green color.}
	\label{network}
	\vspace{-10pt}
\end{figure*}

\subsection{Extended Object Representation}

In our proposed BPJDet, we train a dense single-stage anchor-based detector to directly predict a set of objects $\{\widehat{\mathcal{O}}\!\in\!\widehat{\mathbf{O}}\!\parallel\!\widehat{\mathcal{O}}\!=\!cat(\widehat{\mathcal{O}}^{box},\widehat{\mathcal{O}}^{dis}), \widehat{\mathbf{O}}\!=\!\widehat{\mathbf{O}}^b \cup \widehat{\mathbf{O}}^p\}$, which contains human body set $\widehat{\mathbf{O}}^b$ and body-part set $\widehat{\mathbf{O}}^p$ concurrently. A typical prediction $\widehat{\mathcal{O}}$ is concatenated of the bounding box $\widehat{\mathcal{O}}^{box}$ and corresponding center point displacement $\widehat{\mathcal{O}}^{dis}$. It locates an object with a tight bounding box $\mathbf{\hat{b}}\!=\!(\hat{b}_x, \hat{b}_y, \hat{b}_w, \hat{b}_h)$ where coordinates $(\hat{b}_x, \hat{b}_y)$ are the center position, $\hat{b}_w$ and $\hat{b}_h$ are the width and height of $\mathbf{\hat{b}}$, respectively. It also records the relative displacement $\mathbf{\hat{d}}\!=\!(\hat{d}_{x_i}, \hat{d}_{y_i})|^k_{i=1}$ of center point of body-part ($\widehat{\mathcal{O}}\!\in\!\widehat{\mathbf{O}}^b$) or affiliated body ($\widehat{\mathcal{O}}\!\in\!\widehat{\mathbf{O}}^p$) to each $\mathbf{\hat{b}}$. Considering that body-part may be more than one component (e.g., both hands part with $k=2$), we allow $\mathbf{\hat{d}}$ to be compatible with $k$-dimensional 2D offsets. For these regressed offsets, we will explain how to decode them for the body-part association with fusing detected body-part set $\widehat{\mathbf{O}}^p$ in Section \ref{inference}.

Intuitively, we can benefit a lot from this extended representation. On one hand, an appropriate large body bounding box $\widehat{\mathcal{O}}^{box}$ possesses both strong local characteristics and weak global features (such as surrounding background and anatomical position) for its body part $\widehat{\mathcal{O}}^{dis}$ offset regression. This enables the network to learn their intrinsic relationships. On the other hand, compared to methods \cite{chi2020relational, wan2021body, narasimhaswamy2022whose} training multiple subnetworks or stages, mixing $\widehat{\mathcal{O}}^{box}$ and $\widehat{\mathcal{O}}^{dis}$ up can be leveraged directly in BPJDet without the need of complicated post-processing. By designing a one-stage network that uses shared heads to jointly predict $\widehat{\mathcal{O}}^{box}$ and $\widehat{\mathcal{O}}^{dis}$, our approach can achieve high accuracy with minimal computational burden during training and inference.

\vspace{-5pt}
\subsection{Overall Network Architecture}\label{netarch}

Our network structure is shown in Fig.~\ref{network} left. We choose the recently most cost-effective one-stage YOLOv5 \cite{jocher2020ultralytics} as the basic backbone $\mathcal{N}$. Specifically, following YOLOv5, we feed $\mathcal{N}$ one RGB image $\mathbf{I}\!\in\!\mathbb{R}^{h\times w\times 3}$ as the input, keep its beneficial data augmentation strategies (e.g., Mosaic and MixUp), and output four grids $\widehat{\mathit{G}}\!=\!\{\widehat{\mathcal{G}}^s\!\parallel\!s\!\in\!\{8,16,32,64\}\}$ from four multi-scale heads. Each grid $\widehat{\mathcal{G}}\!\in\!\mathbb{R}^{A_a\times A_o\times \frac{h}{s}\times \frac{w}{s}}$ contains dense object outputs $\widehat{\mathbf{O}}$ produced from $A_a$ anchor channels and $A_o$ output channels. Supposing that one target object $\mathcal{O}$ is centered at $(b_x, b_y)$ in the feature map $\mathbf{F}^s$, the corresponding grid $\widehat{\mathcal{G}}^s$ at cell $(\frac{b_x}{s}, \frac{b_y}{s})$ should be highly confident. When having defined $A_a$ anchor boxes $\mathcal{B}^s\!=\!\{(B^w_i,  B^h_i)\vert^{A_a}_i\}$ for the grid $\widehat{\mathcal{G}}^s$, we will generate $A_a$ anchor channels at each cell $(\frac{b_x}{s}, \frac{b_y}{s})$. Furthermore, to obtain robust capability, YOLOv5 allows detection redundancy of multiple objects and four surrounding cells matching for each cell. This redundancy makes sense to the detection of body or part position $\mathcal{O}^{box}$, but it is not completely facilitative to the regression of offsets $\mathcal{O}^{dis}$. We interpret this in Section \ref{loss}.

Then, we explain the arrangement of one prediction $\widehat{\mathcal{O}}$ from $A_o$ output channels of $\widehat{\mathcal{G}}^s_{i,x,y}$ which is related to $i$-th anchor box at grid cell $(x,y)$. As shown in the example of grid cells in Fig.~\ref{network} right, one typical predicted $\widehat{\mathcal{O}}$ embedding consists of four parts: the objectness or probability $\hat{o}$ that an object exists, the candidate bounding box $\mathbf{\hat{b}}'=({\hat{b}_x}', {\hat{b}_y}', {\hat{b}_w}', {\hat{b}_h}')$, the object classification score $\hat{c}_{k+1}$, and the candidate of body-part offsets $\mathbf{\hat{d}}'\!=\!(\hat{d}'_{x_i}, \hat{d}'_{y_i})|^k_{i=1}$. Thus, $A_o\!=\!3k+6$. To transform the candidate $\mathbf{\hat{b}}'$ into coordinates $\mathbf{\hat{b}}$ relative to the grid cell $\widehat{\mathcal{G}}^s_{i,x,y}$, we apply conversions as below:
\begin{equation}\small
\begin{aligned}
	&\enspace{\hat{b}_x}=2\phi({\hat{b}_x}')-0.5, \qquad {\hat{b}_y}=2\phi({\hat{b}_y}')-0.5 ~\\
	&{\hat{b}_w}=\frac{B^w_i}{s}[2\phi({\hat{b}_w}')]^2, \qquad {\hat{b}_h}=\frac{B^h_i}{s}[2\phi({\hat{b}_h}')]^2 ~
\end{aligned}
\end{equation}
where $\phi$ is the sigmoid function that limits model predictions in the range $(0,1)$. Similarly, this detection strategy can be extended to the offset of body-part. A body-part’s intermediate offsets $\mathbf{\hat{d}}'$ are predicted in the grid coordinates and relative to the grid cell origin $(x,y)$ using:
\begin{equation}\small
	{\hat{d}_{x_i}}\!=\!\frac{B^w_i}{s}[4\phi({\hat{d}_{x_i}}')-2], \:\:\:\:\:\:\:\: \\
	{\hat{d}_{y_i}}\!=\!\frac{B^h_i}{s}[4\phi({\hat{d}_{y_i}}')-2] ~
\end{equation}
In the way, $\hat{d}_{x_i}$ and $\hat{d}_{y_i}$ are constrained to $\pm2\frac{B^w_i}{s}$ and $\pm2\frac{B^h_i}{s}$, respectively. To learn $\mathbf{\hat{b}}$ and $\mathbf{\hat{d}}$, losses are applied in the grid space. Sample targets of $\mathbf{b}$ and $\mathbf{d}$ are shown in Fig.~\ref{network} right.

\vspace{-5pt}
\subsection{Multi-Loss Functions}\label{loss}

For a set of predicted grids $\mathit{\widehat{G}}$, we firstly build target grids set $\mathit{G}$ following formats introduced in Section \ref{netarch}. Then, we mainly compute following four loss components:
\begin{align}\small
	& \mathcal{L}_{box}=\sum\nolimits_s\frac{1}{\|\mathcal{G}^s\|}\sum\nolimits^{\|\mathcal{G}^s\|}_{i=1}[1-\mathsf{CIoU}(\mathbf{\hat{b}_i}, \mathbf{b_i})] \\
	& \mathcal{L}_{obj}=\sum\nolimits_s\frac{w_s}{\|\mathcal{G}^s\|}\sum\nolimits^{\|\mathcal{G}^s\|}_{i=1}\mathsf{BCE}(\hat{o}, o\cdot\mathsf{CIoU}(\mathbf{\hat{b}_i}, \mathbf{b_i})) \\
	& \mathcal{L}_{cls}=\sum\nolimits_s\frac{1}{\|\mathcal{G}^s\|}\sum\nolimits^{\|\mathcal{G}^s\|}_{i=1}\mathsf{BCE}(\hat{c}, c) \\
	& \mathcal{L}_{bpd}=\sum\nolimits_s\frac{1}{\|\mathcal{G}^s\|}\sum\nolimits^{\|\mathcal{G}^s\|}_{i=1}\sum\nolimits^{k}_{1}\varphi(v>0)\| \mathbf{\hat{d}'_i}-\mathbf{d'_i} \|_2 ~
\end{align}

For the bounding box regression loss $\mathcal{L}_{box}$, we adopt the complete intersection over union (CIoU) across four grids $\mathcal{G}^s$. $\mathsf{BCE}$ in the objectness loss $\mathcal{L}_{obj}$ and classification loss $\mathcal{L}_{cls}$ is the binary cross-entropy. The multiplier $o$ in $\mathcal{L}_{obj}$ is used for penalizing candidates without hitting target grid cells ($o=0$), and encouraging candidates around target anchor ground-truths ($o=1$). The $w_s$ is a balance weight for different grid level. Finally, we utilize the mean squared error (MSE) to measure offset outputs $\mathbf{\hat{d}'}$ and normalized targets $\mathbf{d'}$ in body-part displacement loss $\mathcal{L}_{bpd}$. Before that, we apply a filter $\varphi(\cdot)$ with the visibility label of body-part to remove out those false-positive offset predictions from $\widehat{\mathcal{O}}$. Finally, we calculate the total training loss $\mathcal{L}=(\mathit{\widehat{G}},\mathit{G})$ as follows:
\begin{equation}\small
	\mathcal{L}=N_{bs}(\alpha\mathcal{L}_{box} + \beta\mathcal{L}_{obj} + \gamma\mathcal{L}_{cls} + \lambda\mathcal{L}_{bpd}) ~
	\label{losstotal}
\end{equation}
where $N_{bs}$ is the batch size. The $\alpha$, $\beta$, $\gamma$ and $\lambda$ are weights of losses $\mathcal{L}_{box}$, $\mathcal{L}_{obj}$, $\mathcal{L}_{cls}$ and $\mathcal{L}_{bpd}$, respectively.


\subsection{Association Decoding}\label{inference}

In inference, we need to process the predicted objects set $\widehat{\mathbf{O}}$ to get final results. First of all, we apply the conventional Non-Maximum Suppression (NMS) to filter out false-positive and redundant bounding boxes of both body and part objects:
\begin{equation}\small
	\widehat{\mathbf{O}}^{b'}\!= \mathsf{NMS}(\widehat{\mathbf{O}}^b, \tau^b_{conf}, \tau^b_{iou}), \:\:\:\: \\
	\widehat{\mathbf{O}}^{p'}\!= \mathsf{NMS}(\widehat{\mathbf{O}}^p, \tau^p_{conf}, \tau^p_{iou})~
\end{equation}
where $\tau_{conf}$ and $\tau_{iou}$ are thresholds for object confidence and IoU overlap, respectively. We fetch confidence of each predicted object $\widehat{\mathcal{O}}$ by $\hat{o}\cdot\hat{c}_i$, where $i\!=\!1$ for body object and $i\!>\!1$ for part object. Then, we rescale $\widehat{\mathcal{O}}^{box'}$ and $\widehat{\mathcal{O}}^{dis'}$ in $\widehat{\mathbf{O}}^{b'}$ and $\widehat{\mathbf{O}}^{p'}$ to obtain real $\widetilde{\mathbf{O}}^{b}$ and $\widetilde{\mathbf{O}}^{p}$ by below transformations:
\begin{equation}\small
	\widetilde{\mathcal{O}}^{box}\!=\!s\cdot[\widehat{\mathcal{O}}^{box'}+(x_o, y_o, 0, 0)], \:\:\:\: \\
	\widetilde{\mathcal{O}}^{dis}\!=\!s\cdot[\widehat{\mathcal{O}}^{dis'}+(x_o, y_o)] ~
\end{equation}
where $x_o$ and $y_o$ are offsets from grid cell centers. The $s$ maps the box and offset size back to the original image shape.

Finally, we update associated body parts of each left body object in $\widetilde{\mathbf{O}}^{b}$ by fusing its regressed center point offsets ($\widetilde{\mathcal{O}}^{dis}\!\in\!\widehat{\mathbf{O}}^{b'}$) with the remaining candidate part objects $\widetilde{\mathbf{O}}^{p}$. Specifically, we search each box $\widetilde{\mathcal{O}}^{box}$ in part objects with its nearest body-part offset $\widetilde{\mathcal{O}}^{dis}$ belonging to body object. The part box that has a large inner IoU ($>\tau^{inner}_{iou}$) with the body box will be selected. The updated body objects set is denoted as $\widetilde{\mathbf{O}}^{b'}$. We report all evaluation results on $\widetilde{\mathbf{O}}^{b'}$ and $\widetilde{\mathbf{O}}^{p}$.

\section{Experimental Results}

\subsection{Experiment and Training Settings}

{\bf Datasets:} We mainly expect to evaluate the association quality of BPJDet, while maintaining high object detection accuracy. Three public datasets including CityPersons \cite{zhang2017citypersons}, CrowdHuman \cite{shao2018crowdhuman} and BodyHands \cite{narasimhaswamy2022whose} are chosen. The former two are for pedestrian detection tasks. In CityPersons, it has 2,975 and 500 images for training and validation, respectively. In CrowdHuman, there are 15,000 images for training and 4,375 images for validation. Following BFJDet \cite{wan2021body}, we use its re-annotated box labels of visible faces for conducting corresponding experiments. The last dataset BodyHands is for hand-body associations tasks. It has 18,861 and 1,629 images in train-set and test-set with annotations for hand and body locations and correspondences. We implement body-hand joint detection task in it for comparing.

{\bf Evaluation Metric:} For evaluation metrics, we report the standard VOC Average Precision (AP) metric with IoU=0.5 for object detection of body, face and hands. We also present the log-average miss rate (MR$^{-2}$) of body and its parts. For the association quality of BPJDet, we report the log-average miss matching rate (mMR$^{-2}$) as proposed by BFJDet \cite{wan2021body} of body-face pairs, and present Conditional Accuracy and Joint AP defined by BodyHands \cite{narasimhaswamy2022whose} of body-hand pairs.

{\bf Implementation Details:} We adopt the PyTorch 1.10 and 4 RTX-3090 GPUs for training. Depending on the dataset complexity and scale, we train on the CityPersons, CrowdHuman and BodyHands datasets for 100, 150 and 100 epochs using the SGD optimizer, respectively. Following the original YOLOv5 \cite{jocher2020ultralytics} architecture, we train three kinds of models including BFJDet-S/M/L by controlling the depth and width of bottlenecks in $\mathcal{N}$. The shape of input images is resized and zero-padded to $1536\times1536\times 3$ following settings in JointDet \cite{chi2020relational} and BFJDet \cite{wan2021body}. We adopt the data training augmentation but leave out test time augmentation (TTA). As for many hyperparameters, we keep most of them unchanged, including adaptive anchors boxes $\mathcal{B}^s$, the grid balance weight $w_s$, and the loss weights $\alpha\!=\!0.05$, $\beta\!=\!0.7$ and $\gamma\!=\!0.3$. We set $\lambda\!=\!0.015$ based on ablation studies. When testing, we use thresholds $\tau^b_{conf}\!=\!0.05$, $\tau^b_{iou}\!=\!0.6$, $\tau^p_{conf}\!=\!0.1$, $\tau^p_{iou}\!=\!0.3$ and $\tau^{inner}_{iou}\!=\!0.6$ for applying $\mathsf{NMS}$ on $\widehat{\mathbf{O}}$.

\begin{figure}
	\centering
	\pgfplotsset{compat = newest}
	\pgfplotstableread{./figures/lambda.dat}{\table}
	\begin{tikzpicture}\scriptsize 
	\begin{axis}[
    	xmin = 0.4, xmax = 3.1, ymin = 25, ymax = 32, ylabel = MR$^{-2}$, 
    	xtick distance = 0.5, ytick distance = 1, axis y line*=left, grid = both, 
    	minor tick num = 1, major grid style = {lightgray}, minor grid style = {lightgray!45},
   	width = 0.95\columnwidth, height = 0.45\columnwidth,
   	legend cell align = {left}, legend style={at={(axis cs:2.1, 25.7)}, anchor=south west, fill=none, draw=none}
    	]
	\addplot[red, thick, mark = x, mark size = 3pt] table [x ={X}, y = {MRbody}] {\table};
	\addplot[green, thick, mark = x, mark size = 3pt] table [x ={X}, y = {MRface}] {\table};
	\legend{ {\bf MR$^{-2}$body$\downarrow$}, {\bf MR$^{-2}$face$\downarrow$} };
	\end{axis}
	
	\begin{axis}[
    	xmin = 0.4, xmax = 3.1, ymin = 47.5, ymax = 54.5, ylabel = {\color{teal}{avg-mMR$^{-2}$}}, 
    	xtick distance = 0.5, ytick distance = 1, ylabel near ticks, yticklabel pos=right,
   	width = 0.95\columnwidth, height = 0.45\columnwidth,
   	legend cell align = {left}, legend style={at={(axis cs:2.1, 47.2)}, anchor=south west, fill=none, draw=none}
    	]
	\addplot[teal, very thick, mark = square*, mark size = 2pt] table [x = {X}, y = {mMR}] {\table};
	\legend{ {\bf avg-mMR$^{-2}\downarrow$} };
	\end{axis}
	\end{tikzpicture}
	\caption{The influence of parameter $\lambda$ (x-axis, enlarged 100$\times$).}
	\label{lambda}
	\vspace{-15pt}
\end{figure}
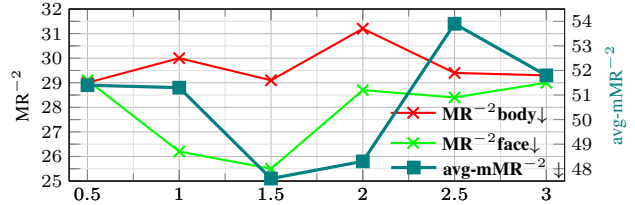 


\setlength{\tabcolsep}{3.8pt} 
\begin{table}[t]\scriptsize 
	\begin{center}
	\caption{The performance comparison of the joint body-face detection task in the val-set of CityPersons.}
	\label{CityPersons}
	\begin{tabular}{l|c|c|cccc}
	\Xhline{1.2pt}
	\multirow{2}{*}{Methods} & \multirow{2}{*}{\makecell{AP$\uparrow$\\body}} & \multirow{2}{*}{\makecell{AP$\uparrow$\\face}} & ~ & mMR$^{-2}\downarrow$ & ~ & ~ \\
	~ & ~ & ~ & {\it Reasonable} & {\it Partial} & {\it Bare} & {\it Heavy} \\
	\Xhline{1.2pt}
	RetinaNet+POS \cite{lin2017focal, wan2021body} & 78.5 & 35.3 & 40.0 & 42.8 & 38.7 & 67.0 \\
	RetinaNet+BFJ \cite{lin2017focal, wan2021body} & 79.3 & 36.2 & 39.5 & 41.5 & 38.5 & 63.1 \\
	FPN+POS \cite{lin2017feature, wan2021body} & 80.6 & 65.5 & 33.5 & 32.7& 34.1 & 56.6 \\
	FPN+BFJ \cite{lin2017feature, wan2021body} & {\bf 84.4} & {\bf 68.0} & 32.7 & 30.6 & 33.0 & 53.5 \\
	\hline
	BPJDet-S (Ours) & 75.1 & 58.2 & 29.3 & 28.9 & 29.3 & 57.2 \\
	BPJDet-M (Ours) & 76.7 & 58.8 & 27.5 & 31.6 & {\bf 24.9} & 55.8 \\
	BPJDet-L (Ours) & 75.5 & 61.0 & {\bf 26.4} & {\bf 27.7} & 25.5 & {\bf 46.2} \\
	\Xhline{1.2pt}
	\end{tabular}
	\end{center}
	\vspace{-20pt}
\end{table}

\setlength{\tabcolsep}{4pt} 
\begin{table}[t]\scriptsize 
	\begin{center}
	\caption{The performance comparison of the joint body-face detection task in the val-set of CrowdHuman.}
	\label{CrowdHuman}
	\begin{tabular}{l|c|c|c|c|c|c}
	\Xhline{1.2pt}
	\multirow{2}{*}{Methods} & \multirow{2}{*}{Stage} & \multirow{2}{*}{\makecell{MR$^{-2}$\\body$\downarrow$}} & \multirow{2}{*}{\makecell{MR$^{-2}$\\face$\downarrow$}} & \multirow{2}{*}{\makecell{AP$\uparrow$\\body}} & \multirow{2}{*}{\makecell{AP$\uparrow$\\face}} & \multirow{2}{*}{mMR$^{-2}\downarrow$} \\
	~ & ~ & ~ & ~ & ~ & ~ \\
	\Xhline{1.2pt}
	RetinaNet+POS \cite{lin2017focal, wan2021body} & One & 52.3 & 60.1 & 79.6 & 58.0 & 73.7 \\
	RetinaNet+BFJ \cite{lin2017focal, wan2021body} & One & 52.7 & 59.7 & 80.0 & 58.7 & 63.7 \\ 
	FPN+POS \cite{lin2017feature, wan2021body} & Two & 43.5 & 54.3 & 87.8 & 70.3 & 66.0 \\
	FPN+BFJ \cite{lin2017feature, wan2021body} & Two & 43.4 & 53.2 & 88.8 & 70.0 & 52.5 \\
	CrowdDet+POS \cite{chu2020detection, wan2021body} & Two & 41.9 & 54.1 & {\bf 90.7} & 69.6 & 64.5 \\ 
	CrowdDet+BFJ \cite{chu2020detection, wan2021body} & Two & 41.9 & 53.1 & 90.3 & 70.5 & 52.3 \\ 
	\hline
	BPJDet-S (Ours) & One & 41.3 & 45.9 & 89.5 & 80.8 & 51.4 \\
	BPJDet-M (Ours) & One & {\bf 39.7} & {\bf 45.0} & {\bf 90.7} & {\bf 82.2} & 50.6 \\
	BPJDet-L (Ours) & One & 40.7 & 46.3 & 89.5 & 81.6 & {\bf 50.1} \\
	\Xhline{1.2pt}
	\end{tabular}
	\end{center}
	\vspace{-20pt}
\end{table}

\setlength{\tabcolsep}{2pt} 
\begin{table}[t]\scriptsize  
	\begin{center}
	\caption{The performance comparison of the joint body-hand detection task in the val-set of BodyHands. The * means using hand self-association option.}
	\label{BodyHands}
	\begin{tabular}{l|c|c|c}
	\Xhline{1.2pt}
	Methods & Hand AP$\uparrow$ & Cond. Accuracy$\uparrow$ & Joint AP$\uparrow$ \\
	\Xhline{1.2pt}
	OpenPose \cite{cao2017realtime} & 39.7 & 74.03 & 27.81 \\
	Keypoint Communities \cite{zauss2021keypoint} & 33.6 & 71.48 & 20.71 \\
	MaskRCNN+Feature Distance \cite{he2017mask, narasimhaswamy2022whose} & 84.8 & 41.38 & 23.16 \\
	MaskRCNN+Feature Similarity \cite{he2017mask, narasimhaswamy2022whose} & 84.8 & 39.12 & 23.30 \\
	MaskRCNN+Locaction Distance \cite{he2017mask, narasimhaswamy2022whose} & 84.8 & 72.83 & 50.42 \\
	MaskRCNN+IoU \cite{he2017mask, narasimhaswamy2022whose} & 84.8 & 74.52 & 51.74 \\
	BodyHands \cite{narasimhaswamy2022whose} & 84.8 & 83.44 & 63.48 \\
	BodyHands* \cite{narasimhaswamy2022whose} & 84.8 & 84.12 & 63.87 \\
	\hline
	BPJDet-S (Ours) & 84.0 & 85.68 & 77.86 \\
	BPJDet-M (Ours) & 85.3 & 86.80 & 78.13 \\
	BPJDet-L (Ours) & {\bf 85.9} & {\bf 86.91} & {\bf 84.39} \\
	\Xhline{1.2pt}
	\end{tabular}
	\end{center}
	\vspace{-20pt}
\end{table}

\subsection{Quantitative and Visual Comparison}

{\bf Ablation Studies:} We mainly investigate the hyperparameter $\lambda$. For simplicity, we conduct all ablation experiments on the CityPersons dataset using BPJDet-S and training about joint body-face detection task. The input shape is $1280\times1280\times3$. Total epoch is expanded to 150 for searching the best result. We uniformly sample $\lambda$ from 0.005 to 0.030 with step 0.005 for training, and report metrics including MR$^{-2}$s and average mMR$^{-2}$ of each best model. As in Fig.~\ref{lambda}, we can clearly find that the model performance is optimal when $\lambda\!=\!0.015$. A larger or smaller $\lambda$ will lead to inferior results. More ablation studies are provided in our supplementary material.

{\bf Results on CityPersons and CrowdHuman:} We compare joint body-face detection performance of BPJDet with method BFJDet \cite{wan2021body} on these two benchmarks. Table \ref{CityPersons} shows results on the val-set of CityPersons. Following \cite{wang2018repulsion, wan2021body}, results are reported on four subsets of {\it Reasonable} (occlusion $<$ 35\%), {\it Partial} (10\% $<$ occlusion $\leqslant$ 35\%), {\it Bare} (occlusion $\leqslant$ 10\%) and {\it Heavy} (occlusion $>$ 35\%). Comparing with RetinaNet+BFJ which obtains similar body AP but far inferior face AP to ours, BPJDet-L exceeds it of mMR$^{-2}$ by {\bf 13.1}\%, {\bf 13.8}\%, {\bf 13.0}\% and {\bf 16.9}\% in four subsets, respectively. Comparing with FPN+BFJ that has similar face AP but higher body AP to ours, BPJDet-L also surpasses it of mMR$^{-2}$ by {\bf 6.3}\%, {\bf 2.9}\%, {\bf 7.5}\% and {\bf 7.3}\% in four subsets. Especially, in the {\it Heavy} subset, our association superiority is the most prominent, which reveals that our BPJDet has an advantage for addressing miss-matchings in crowded scenes.

\begin{figure}[]
	\includegraphics[height = 0.33\columnwidth]{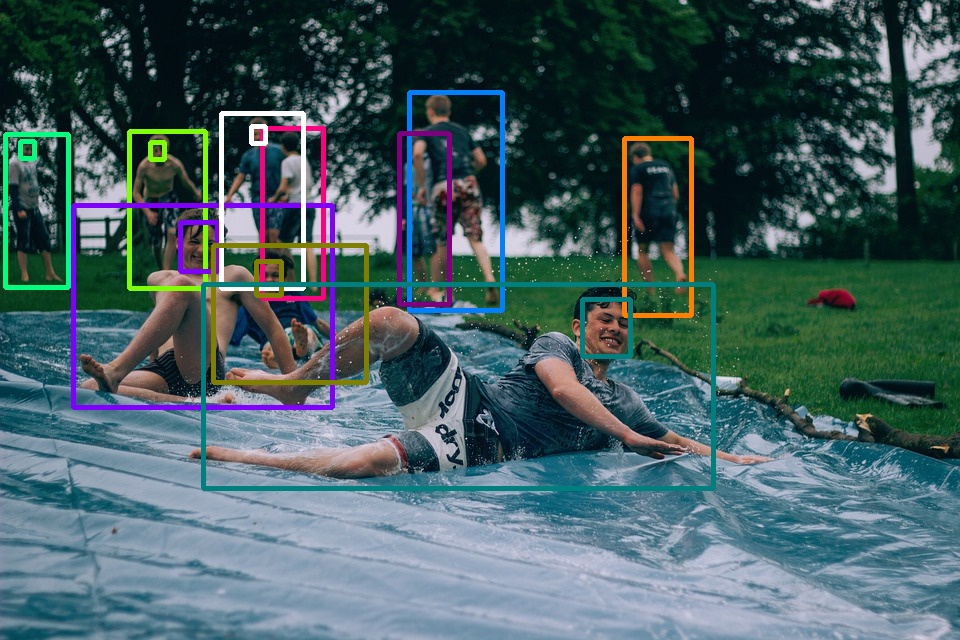}
	\includegraphics[height = 0.33\columnwidth]{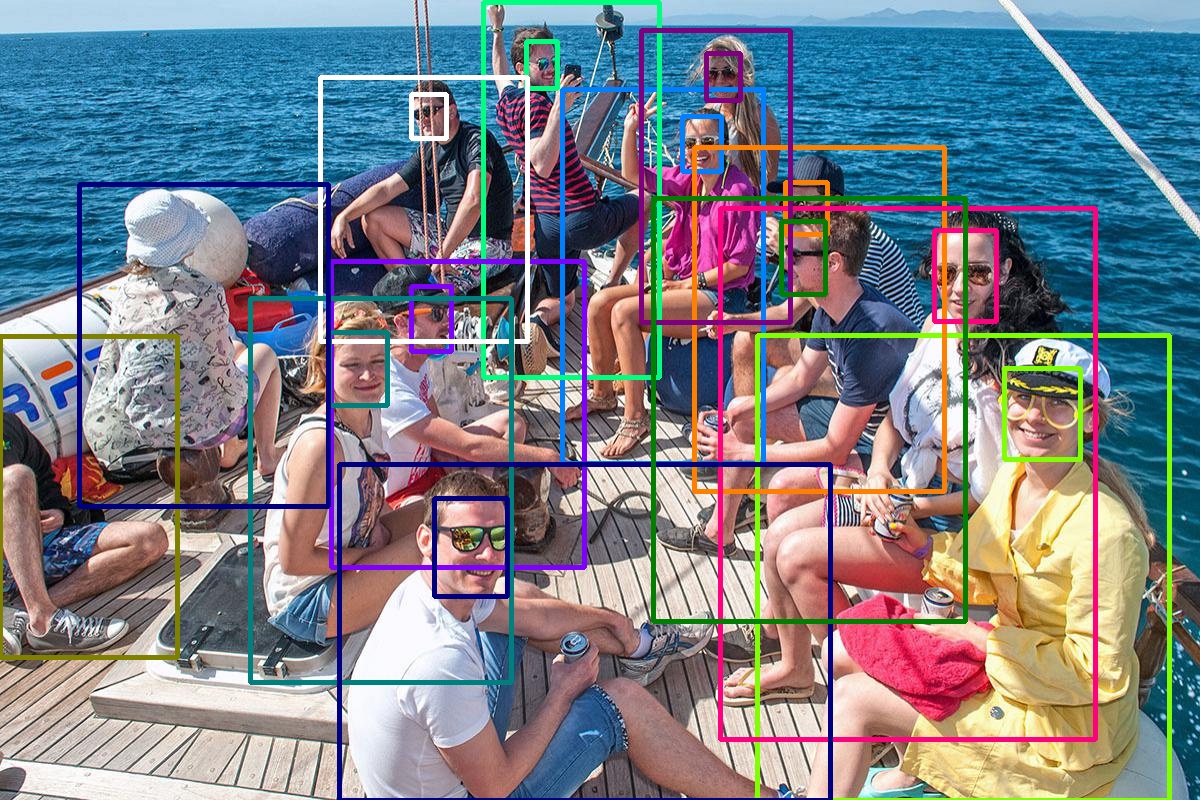}\\
	\includegraphics[height = 0.339\columnwidth]{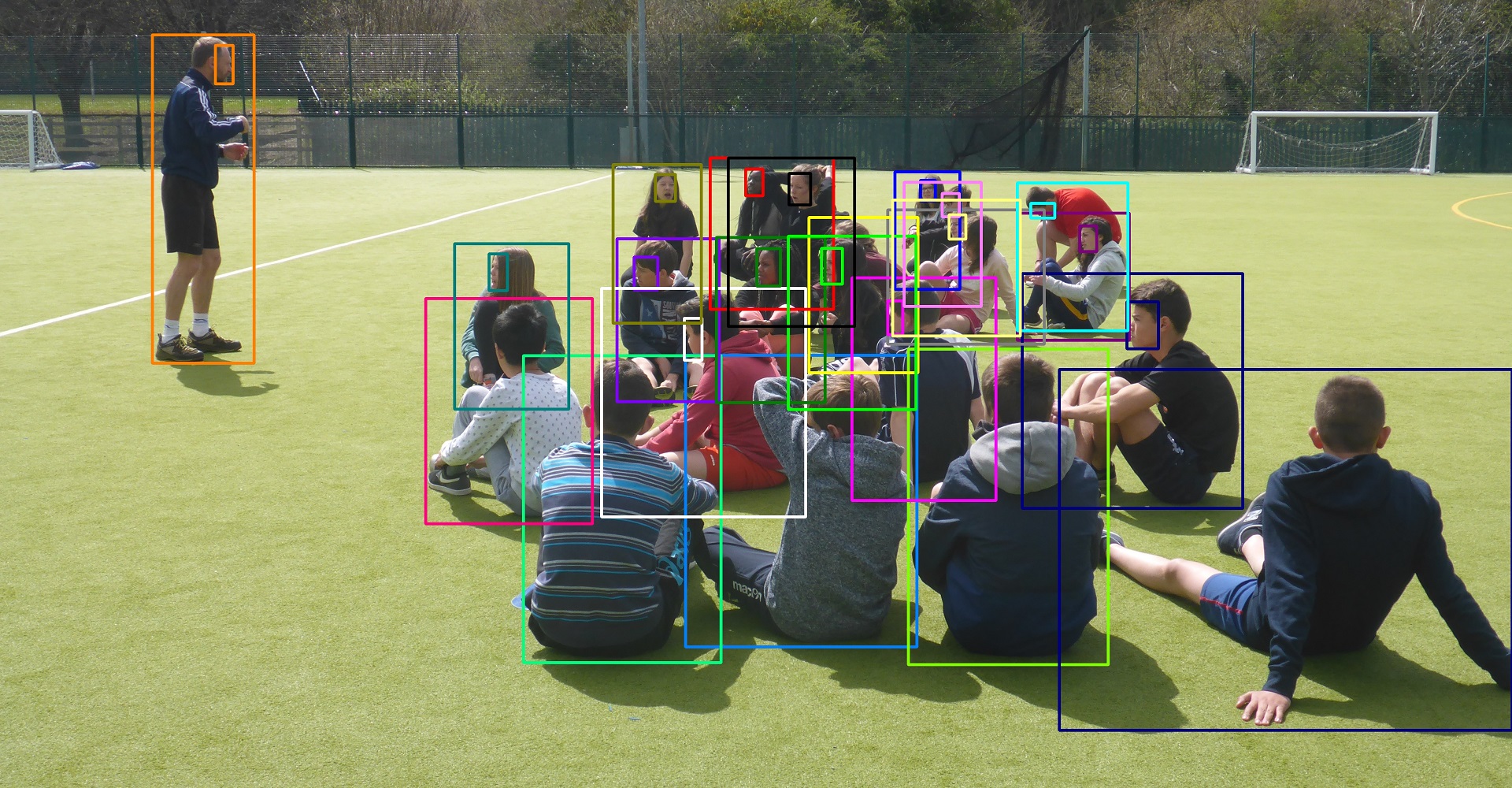}
	\includegraphics[height = 0.339\columnwidth]{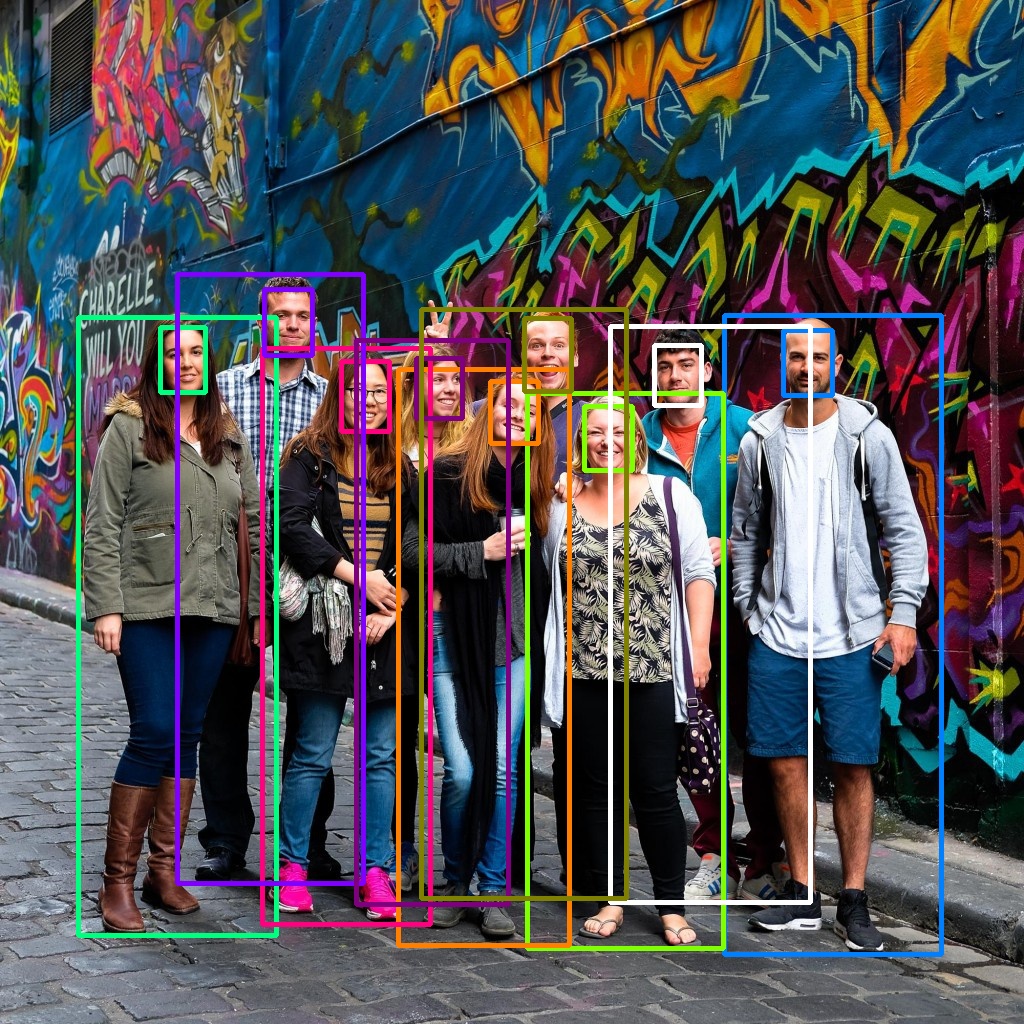}\\ 
	\vspace{-20pt}  
	\caption{Results of BPJDet-L on CrowdHuman val-set images.}
	\label{BPJDetFace}
\end{figure}

Table \ref{CrowdHuman} shows results on the more challenging CityPersons. Our method BPJDet achieves considerable gains in all metrics comparing with BFJDet based on one-stage RetinaNet \cite{lin2017focal} or two-stage FPN \cite{lin2017feature} and CrowdDet \cite{chu2020detection}. Remarkably, BPJDet-L has achieved the lowest mMR$^{-2}$ value {\bf 50.1\%}, which is {\bf 2.2}\% lower than the previous best method CrowdDet+BFJ. These again demonstrate the superiority of our method. More persuasive visual results of BPJDet-L trained on CrowdHuman and tested on complicated overcrowded scenes are shown in Fig. \ref{BPJDetFace}.

\begin{figure}[]
	\includegraphics[height = 0.245\columnwidth]{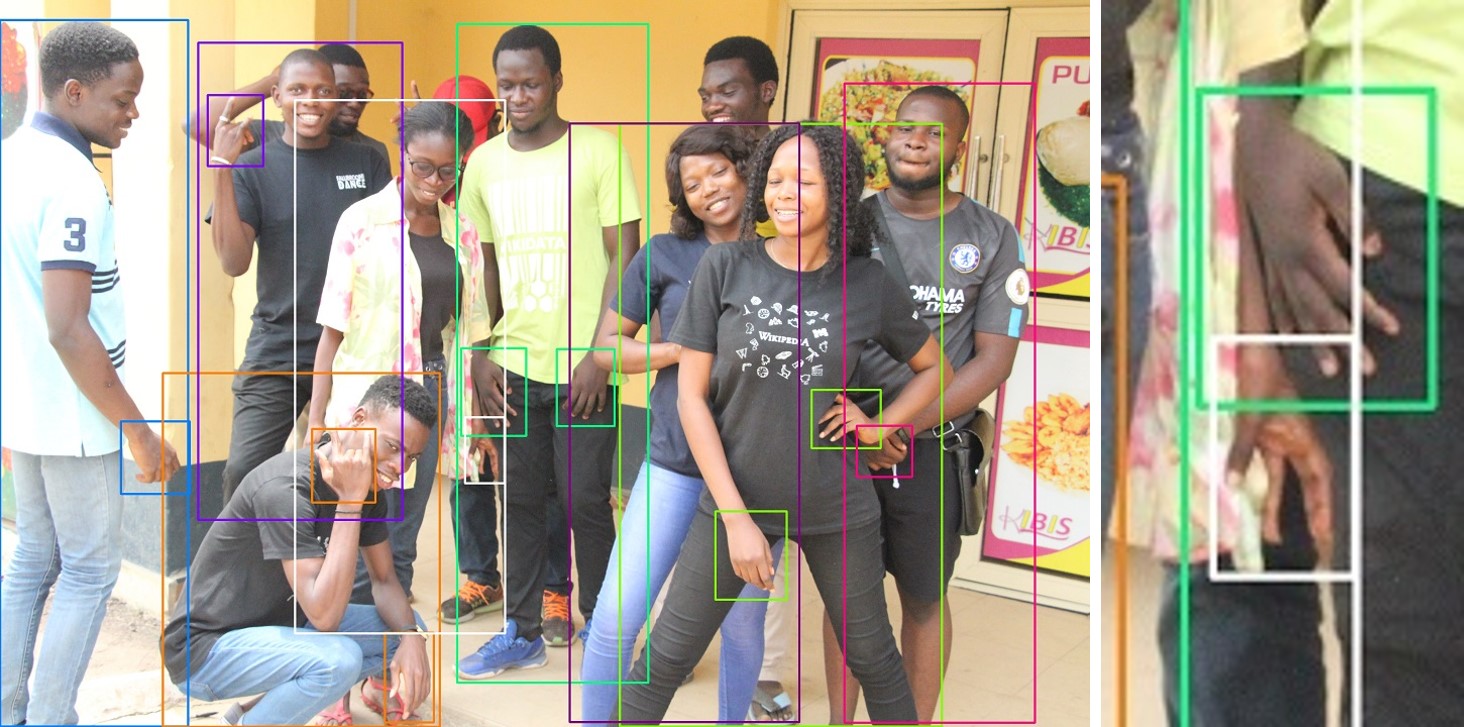}
	\includegraphics[height = 0.245\columnwidth]{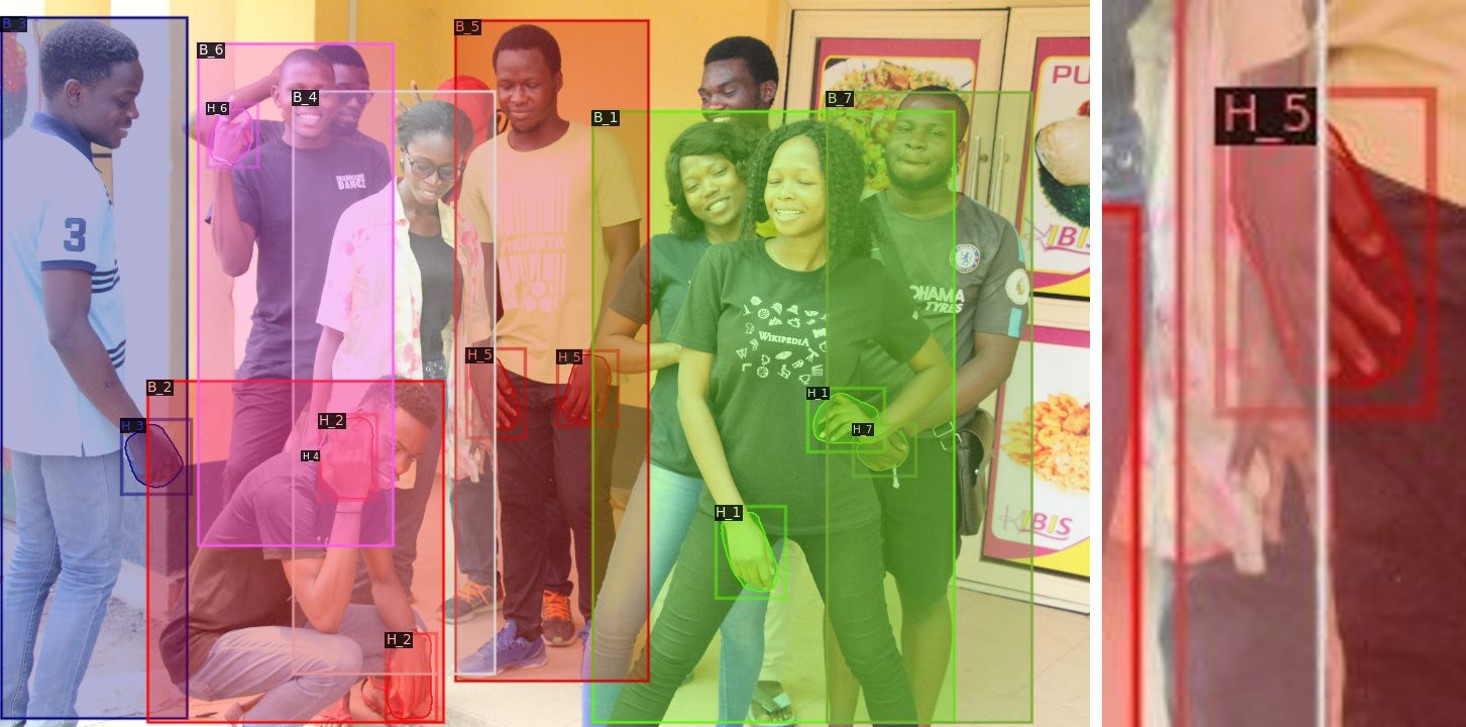}\\
	\includegraphics[height = 0.2335\columnwidth]{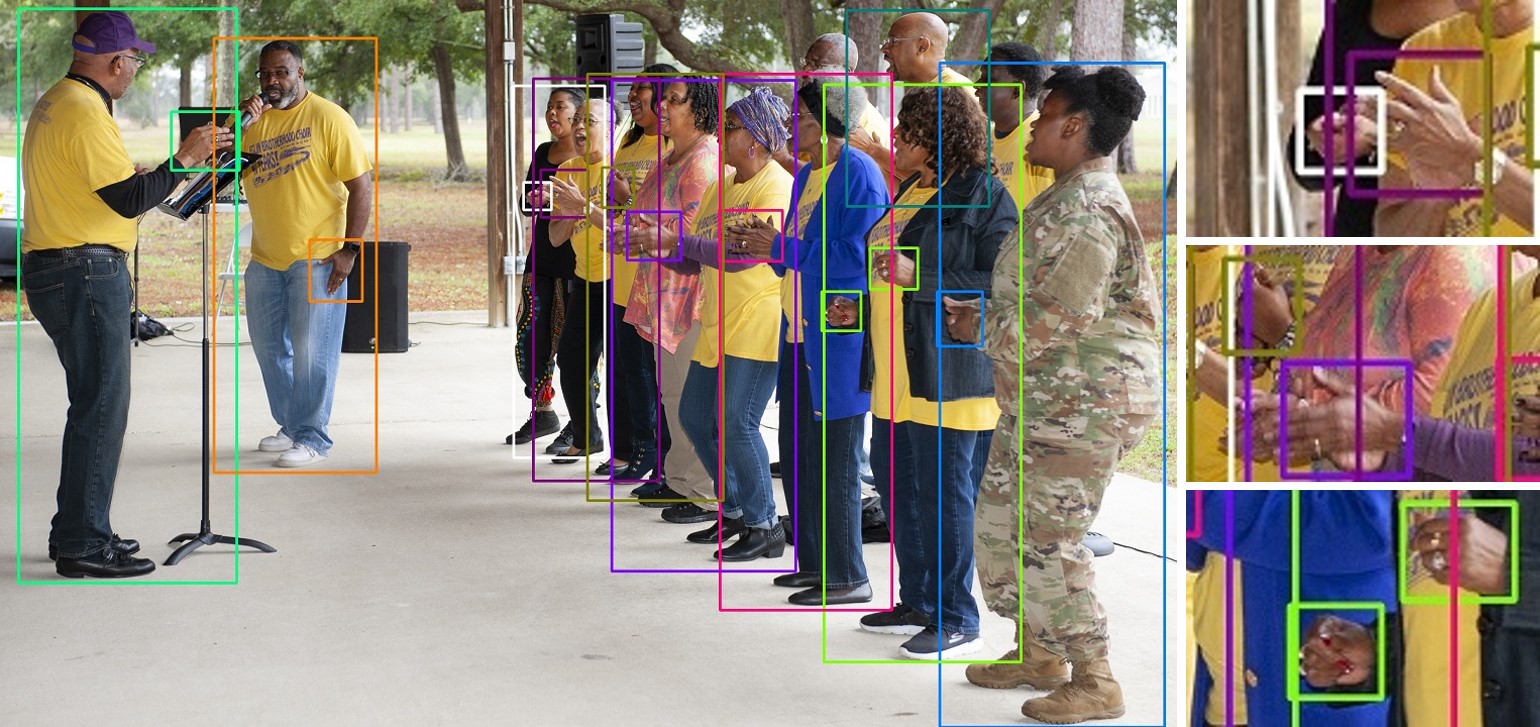}
 	\includegraphics[height = 0.2335\columnwidth]{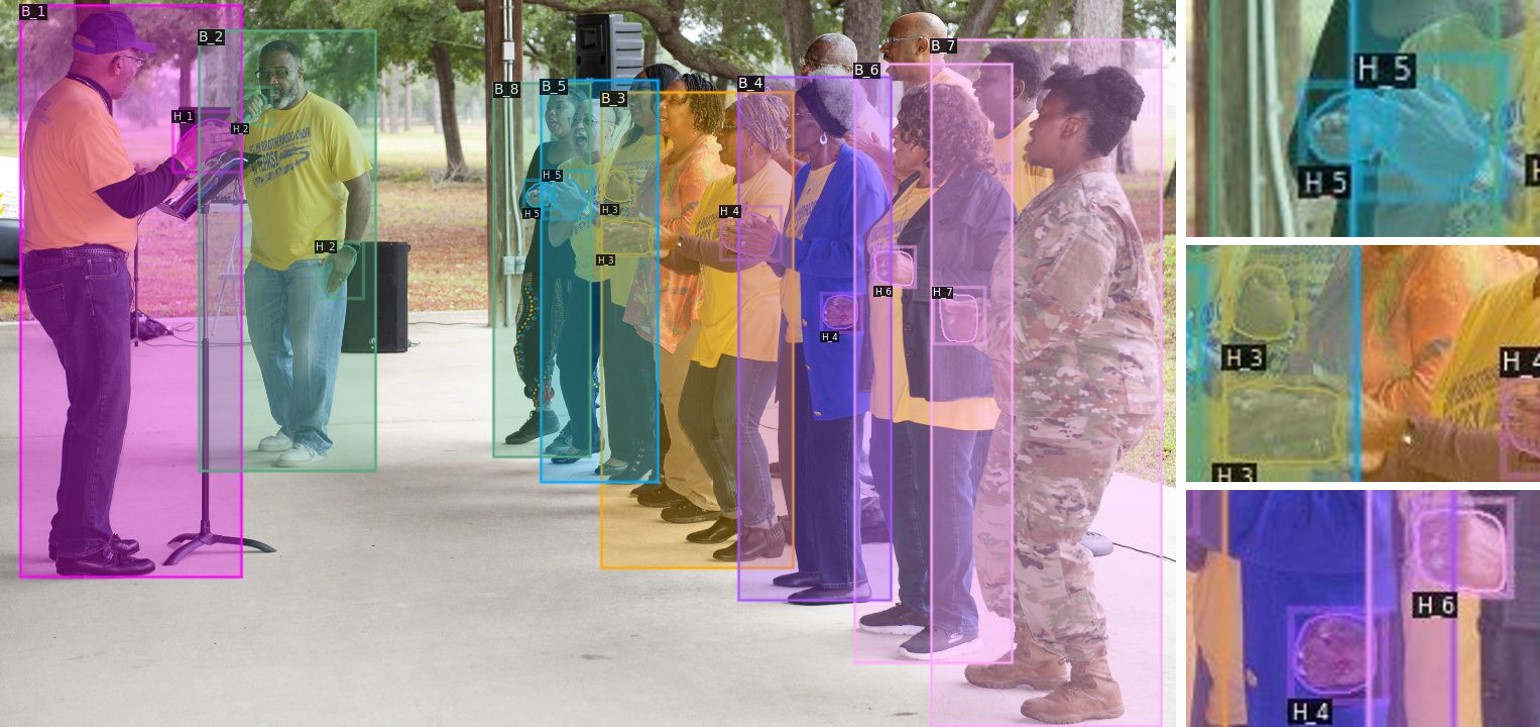}\\
	\includegraphics[height = 0.244\columnwidth]{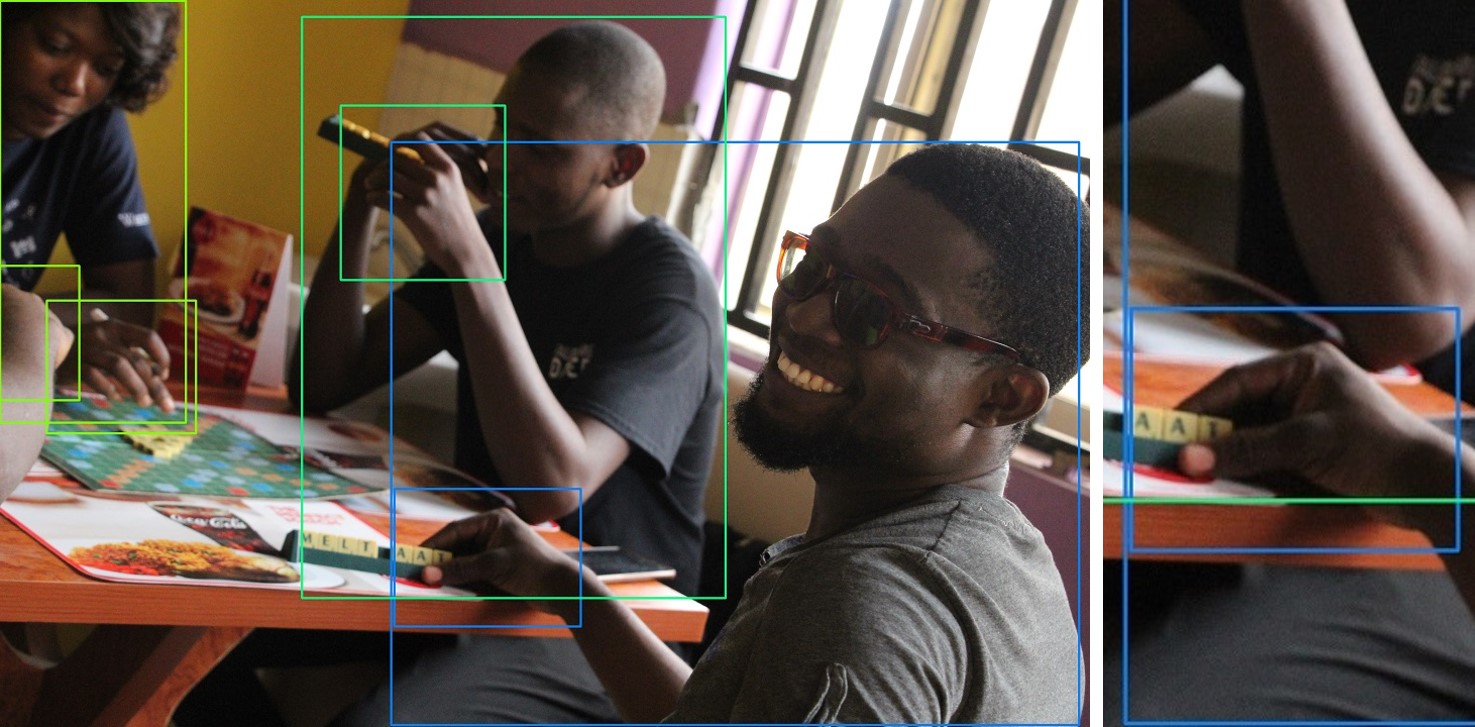}
	\includegraphics[height = 0.244\columnwidth]{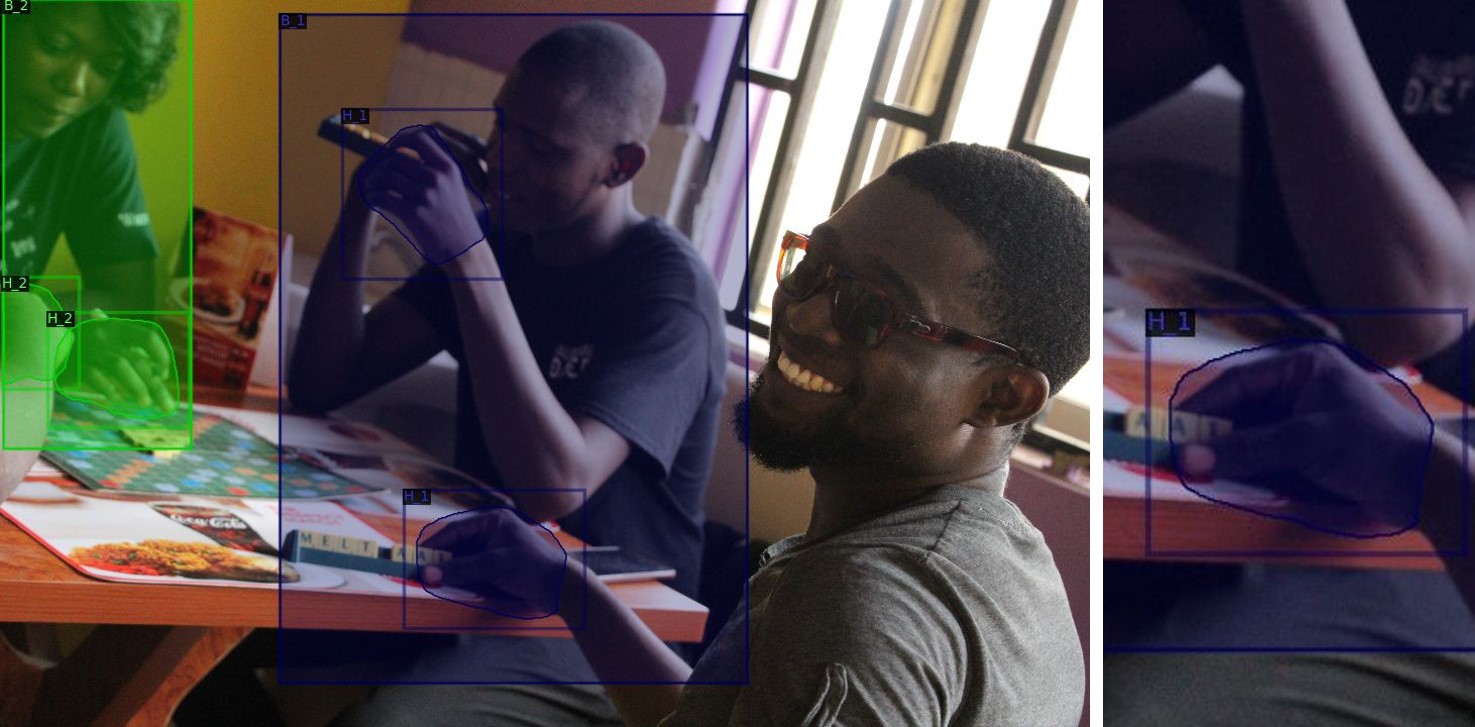}\\
	\vspace{-20pt}  
	\caption{Qualitative results comparison of our BPJDet-L ({\bf left}) with the method in BodyHands ({\bf right}) on its val-set images.}
	\label{BPJDetHand}
\end{figure}

{\bf Results on BodyHands:} We conduct joint body-hand detection experiments on BodyHands \cite{narasimhaswamy2022whose}, and compare our BPJDet with several methods proposed in BodyHands. As shown in Table \ref{BodyHands}, our BPJDet outperforms all other methods by a significant margin. With achieving highest hand AP 85.9\% and conditional accuracy 86.91\% of body, BPJDet-L largely improves the previous best Joint AP of body and hands by {\bf 20.52}\%. We give more qualitative comparisons of joint body-hand detection trained on BodyHands in Fig. \ref{BPJDetHand}. The compared masked images are fetched from BodyHands paper. Our BPJDet can detect many unlabeled objects, and associate hands that method proposed in BodyHands fails to match. This illustrates why we have an overwhelming Joint AP advantage. All these results further validate the generalizability of our extended object representation and its impressive strength on body and part relationship discovery.

\section{Conclusion}

In this paper, we propose a novel body-part joint detector named BPJDet to address the challenging paired object detection and association problem. Inspired by offset regression in general object detection and human pose estimation, we attempt to extend the traditional object representation by appending body-part displacements, and design favorable multi-loss functions to enable joint training. Furthermore, BPJDet is not limited to a specific or single body part. Quantitative SOTA results and impressive qualitative performance on three public datasets suffice to demonstrate the robustness, adaptability and superiority of our proposed BPJDet.



\bibliographystyle{IEEEbib}
{\small \bibliography{refs}}

\end{document}